\documentclass[sigconf]{acmart}
\AtBeginDocument{%
  }

\setcopyright{acmlicensed}
\copyrightyear{2018}
\acmYear{2018}
\acmDOI{XXXXXXX.XXXXXXX}
\acmConference[Conference acronym 'XX]{Make sure to enter the correct
  conference title from your rights confirmation email}{June 03--05,
  2018}{Woodstock, NY}
\acmISBN{978-1-4503-XXXX-X/2018/06}

\begin{document}

\title{Textual and Visual Guided Task Adaptation for Source-Free Cross-Domain Few-Shot Segmentation}

\author{Jianming Liu}
\authornote{The corresponding author.}
\orcid{0009-0007-5755-2729}
\affiliation{%
  \department{School of Artificial Intelligence}
  \institution{Jiangxi Normal University}
  \city{Nanchang}
  \state{Jiangxi}
  \postcode{330022}
  \country{China}
}
\email{liujianming@jxnu.edu.cn}

\author{Wenlong Qiu}
\orcid{0009-0006-8717-9884}
\affiliation{%
  \department{School of Digital Industry}
  \institution{Jiangxi Normal University}
  \city{Shangrao}
  \state{Jiangxi}
  \postcode{334000}
  \country{China}}
\email{202341600192@jxnu.edu.cn}
  
\author{Haitao Wei}
\orcid{0009-0008-6924-6486}
\affiliation{%
  \department{School of Digital Industry}
  \institution{Jiangxi Normal University}
  \city{Shangrao}
  \state{Jiangxi}
  \postcode{334000}
  \country{China}}
\email{202241600194@jxnu.edu.cn}

\renewcommand{\shortauthors}{Jianming Liu,  Wenlong Qiu, \& Haitao Wei}

\begin{abstract}
Few-Shot Segmentation(FSS) aims to efficient segmentation of new objects with few labeled samples. However, its performance significantly degrades when domain discrepancies exist between training and deployment. Cross-Domain Few-Shot Segmentation(CD-FSS) is proposed to mitigate such performance degradation. Current CD-FSS methods primarily sought to develop segmentation models on a source domain capable of cross-domain generalization. However, driven by escalating concerns over data privacy and the imperative to minimize data transfer and training expenses, the development of source-free CD-FSS approaches has become essential. In this work, we propose a source-free CD-FSS method that leverages both textual and visual information to facilitate target domain task adaptation without requiring source domain data. Specifically, we first append Task-Specific Attention Adapters (TSAA) to the feature pyramid of a pretrained backbone, which adapt multi-level features extracted from the shared pre-trained backbone to the target task. Then, the parameters of the TSAA are trained through a Visual-Visual Embedding Alignment (VVEA) module and a Text-Visual Embedding Alignment (TVEA) module. The VVEA module utilizes global-local visual features to align image features across different views, while the TVEA module leverages textual priors from pre-aligned multi-modal features (e.g., from CLIP) to guide cross-modal adaptation. By combining the outputs of these modules through dense comparison operations and subsequent fusion via skip connections, our method produces refined prediction masks. Under both 1-shot and 5-shot settings, the proposed approach achieves average segmentation accuracy improvements of 2.18\% and 4.11\%, respectively, across four cross-domain datasets, significantly outperforming state-of-the-art CD-FSS methods. Code are available at \url{https://github.com/ljm198134/TVGTANet}.
\end{abstract}



\maketitle

\section{Introduction}

\begin{figure}[t]
    \centering
    \includegraphics[width=\columnwidth]{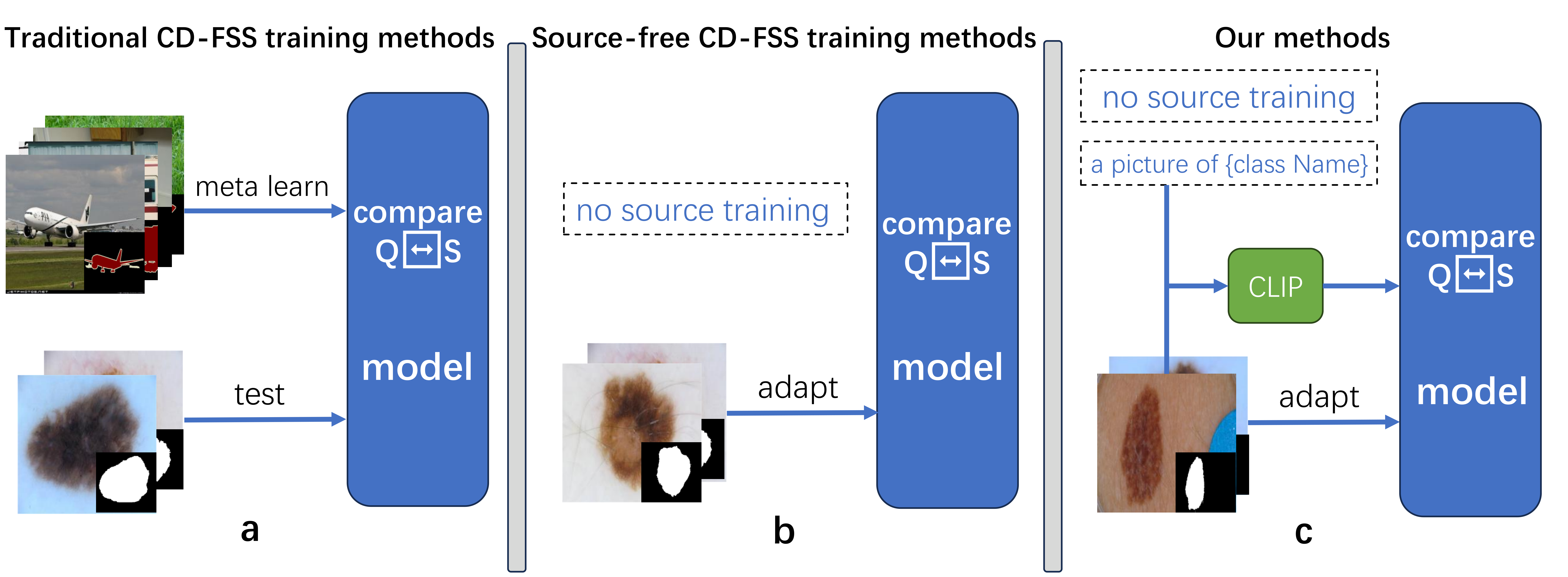}
    \vspace{-15pt} 
    \caption{Architectural Comparison of Cross-Domain Few-Shot Segmentation Methods}
    \Description{A diagram comparing the architectures of different cross-domain few-shot segmentation methods.}
    \label{fig:figure0}
\end{figure}

Few-shot segmentation (FSS)~\cite{1,2,16} aims to generate pixel-level predictions for new classes by using a limited number of support samples; it typically leverages class-agnostic knowledge learned from abundant base classes to facilitate generalization to novel classes~\cite{1,2}. However, most existing FSS methods assume that novel classes and base classes reside in the same domain. Consequently, when applied to cross-domain datasets, their performance degrades significantly. To tackle this challenge, the cross-domain few-shot segmentation (CD-FSS) task~\cite{23} has been proposed, aiming to enhance the cross-domain generalization ability of few-shot segmentation models. Mainstream CD-FSS algorithms~\cite{19,20,21,22,23,25,35} primarily train on a single source domain (e.g., PASCAL VOC 2012~\cite{pascal-voc-2012}), learning domain-invariant features to equip the model with generalization capabilities on the target domain. Their typical architecture and strategy are illustrated in Figure~\ref{fig:figure0}(a). However, these methods are often impractical in many real-world applications. First, in certain application domains (e.g., medical imaging, autonomous driving), source datasets may be inaccessible due to confidentiality, privacy, and copyright issues. Second, training on large-scale source datasets incurs high computational costs, especially for edge devices. Furthermore, meta-training on base datasets with abundant annotated samples inevitably biases the model towards known classes, thereby hindering the recognition of novel concepts~\cite{10}. To address these challenges, a recent study proposed an innovative strategy~\cite{24} that performs segmentation directly on the target domain via a test-time task adaptation approach, without relying on source domain data, achieving promising results, as shown in Figure~\ref{fig:figure0}(b). Nevertheless, this method entirely depends on the limited support samples available in the target domain for model adaptation. Due to the limited number of support samples, this approach is not only prone to overfitting but also substantially weakens the model's ability to capture intra-class appearance diversity in the target domain, making it difficult to fully address the complexity of real-world scenarios. 

To address the aforementioned issues, we reconsider the task adaptation strategy during the test phase and leverage the pre-trained language-image  model CLIP~\cite{31} based on contrastive learning to incorporate class textual information, thereby enhancing the model's generalization capability. CLIP has achieved remarkable success in image-text matching tasks, demonstrating powerful transferability across various datasets~\cite{27,28}. 

In this paper, we introduce a novel source-free cross-domain few-shot segmentation (CD-FSS) method that innovatively leverages both visual and textual information to adapt to target domains without relying on source data. Our approach integrates a Task-Specific Attention Adapter (TSAA), strategically placed within the backbone, with two crucial alignment modules: Visual-Visual Embedding Alignment (VVEA) and Text-Visual Embedding Alignment (TVEA). During training, VVEA ensures robust class-agnostic embedding consistency across augmented views and class-dependent embedding consistency within support sets, while TVEA utilizes tailored text prompts and a frozen CLIP model to generate pseudo segmentation masks for self-supervision. This allows us to train a pixel-level classifier using only the available target domain data. For prediction, our method fuses multi-level coarse similarity maps with refined predictions from both CLIP and the trained pixel-level classifier to generate accurate segmentation masks. By freezing the backbone and CLIP, we focus learning on the task adaptation modules, demonstrating a powerful and data-efficient solution for CD-FSS.

Our contributions are summarized as follows:
\begin{itemize}
    \item We propose TVGTANet, a novel method for source-free cross-domain few-shot semantic segmentation (CD-FSS). This pioneering approach is the first, to our knowledge, to effectively integrate the capabilities of the CLIP model with both textual and visual guidance for robust task adaptation in CD-FSS.
    \item To effectively learn the parameters of the additional adaptation module, we introduce two key alignment modules: VVEA and TVEA. These modules leverage global, local prototype, and cross-modal consistency to significantly enhance class discriminability within the query feature space, while simultaneously preventing overfitting to the support set.
    \item Extensive experiments across four diverse cross-domain datasets demonstrate that TVGTANet achieves superior accuracy and efficiency, outperforming state-of-the-art methods.
\end{itemize}

\section{Related Work}
\subsection{Few-Shot Segmentation}
Few-shot segmentation (FSS) methods predict dense masks for query images using only a small number of annotated support images. Current FSS methods are generally meta-learning-based and can be broadly categorized into two types: prototype-based methods~\cite{1,2,3,4,5,11,12,liu2020part} and matching-based methods~\cite{16,7,8,9}. Prototype-based methods extract class prototypes from support images. They then compute the similarity between support and query images using non-parametric metrics (e.g., cosine similarity)~\cite{10} to guide query image segmentation. To overcome the limitations of single prototypes in pixel-level segmentation, recent research has introduced multi-prototype methods~\cite{2,3,liu2020part}. These methods can represent different parts of an object, thereby improving segmentation accuracy. MIANet~\cite{13},IMR~\cite{14} and PI-Clip~\cite{wang2024rethinking} have also proposed using textual information to assist segmentation, achieving promising results. Matching-based methods~\cite{15,34} connect features of support and query images, employing learnable metrics~\cite{16} to extract dense correspondences between them~\cite{17,18}. By analyzing pixel-level similarities, these methods further optimize query image segmentation results. However, the segmentation performance of these FSS methods degrades rapidly when the domain gap in the target domain is too large.

\subsection{Cross-Domain Few-Shot Segmentation}
Unlike traditional few-shot segmentation, cross-domain few-shot segmentation (CD-FSS) faces a significant domain gap between the source and target domains. CD-FSS requires learning a general segmentation model from a small number of annotated samples in the source domain and effectively transferring it to the target domain. Many methods~\cite{23,21,22,20,19,25} have focused on reducing inter-domain discrepancies and enhancing the learning of domain-invariant features. PATNet~\cite{23} performs feature transformation to map source domain features into domain-invariant representations; however, its inference performance is limited when there is a large intra-class difference between the support and query images. DRANet~\cite{21} introduces a small adapter to align diversified target domain styles with the source domain, thereby improving segmentation performance. DARNet~\cite{22} further reduces the domain gap by perturbing the channel statistics of source domain features and dynamically adjusting matching thresholds, simultaneously enhancing the generalization ability to unknown target domains. APSeg~\cite{19} is based on an encoder-decoder structure that uses SAM as the image encoder to stably transform domain-specific input features into domain-invariant ones; at test time, cross-domain adaptation can be achieved rapidly without fine-tuning the model. Peng et al.~\cite{peng2025sam} propose a SAM-Aware Graph Prompt Reasoning Network for cross-domain few-shot segmentation (CD-FSS). Their approach leverages the powerful segmentation capabilities of SAM, primarily refining its outputs to enhance model adaptability and performance in scenarios with significant domain shifts and limited annotated data. However, achieving these improved results necessitates higher hardware requirements and longer training and interface durations. In contrast to these approaches, ABCDFSS~\cite{24} discards the dependency on source domain data and carries out task adaptation during testing on the target domain. Our work adopts a similar source-free strategy; however, we uniquely introduce textual information to assist in the segmentation of target domain data, substituting the prior knowledge typically extracted from source domain data.

\subsection{Vision-Language Models}

Previous studies~\cite{26,27,28,29,30} have employed textual information to guide models in image segmentation tasks. They have demonstrated the effectiveness of leveraging the strong image-text association capability of CLIP~\cite{31} in segmentation. Among these, clipseg~\cite{29} proposed a segmentation method that combines textual and image prompts, enabling dynamic adaptation to zero-shot (0-shot) and one-shot (1-shot) segmentation tasks based on diverse inputs. DenseCLIP~\cite{30} applies CLIP to dense prediction tasks based on pixel-text matching and combines contextual image information to fully exploit its pre-trained knowledge for improved segmentation performance. Studies~\cite{26,27,28} generate pseudo-labels that integrate both image and textual information by employing CLIP and Grad-CAM~\cite{32}. They rely on these pseudo-labels to enhance model training. In particular, MaskCLIP~\cite{28} generates pseudo-labels in a zero-shot scenario and further improves segmentation performance via self-training. Reference~\cite{26} proposed a text-driven weakly supervised segmentation framework. This framework improves segmentation accuracy through optimized text prompt design and attention mechanisms. Meanwhile, WeCLIP~\cite{27} further refines pseudo-label quality by incorporating a dynamic adjustment mechanism based on~\cite{26}. These methods have all achieved promising results in few-shot segmentation tasks, and we extend their applications to cross-domain few-shot segmentation.
\begin{figure*}[t]
    \centering
    \includegraphics[width=\textwidth]{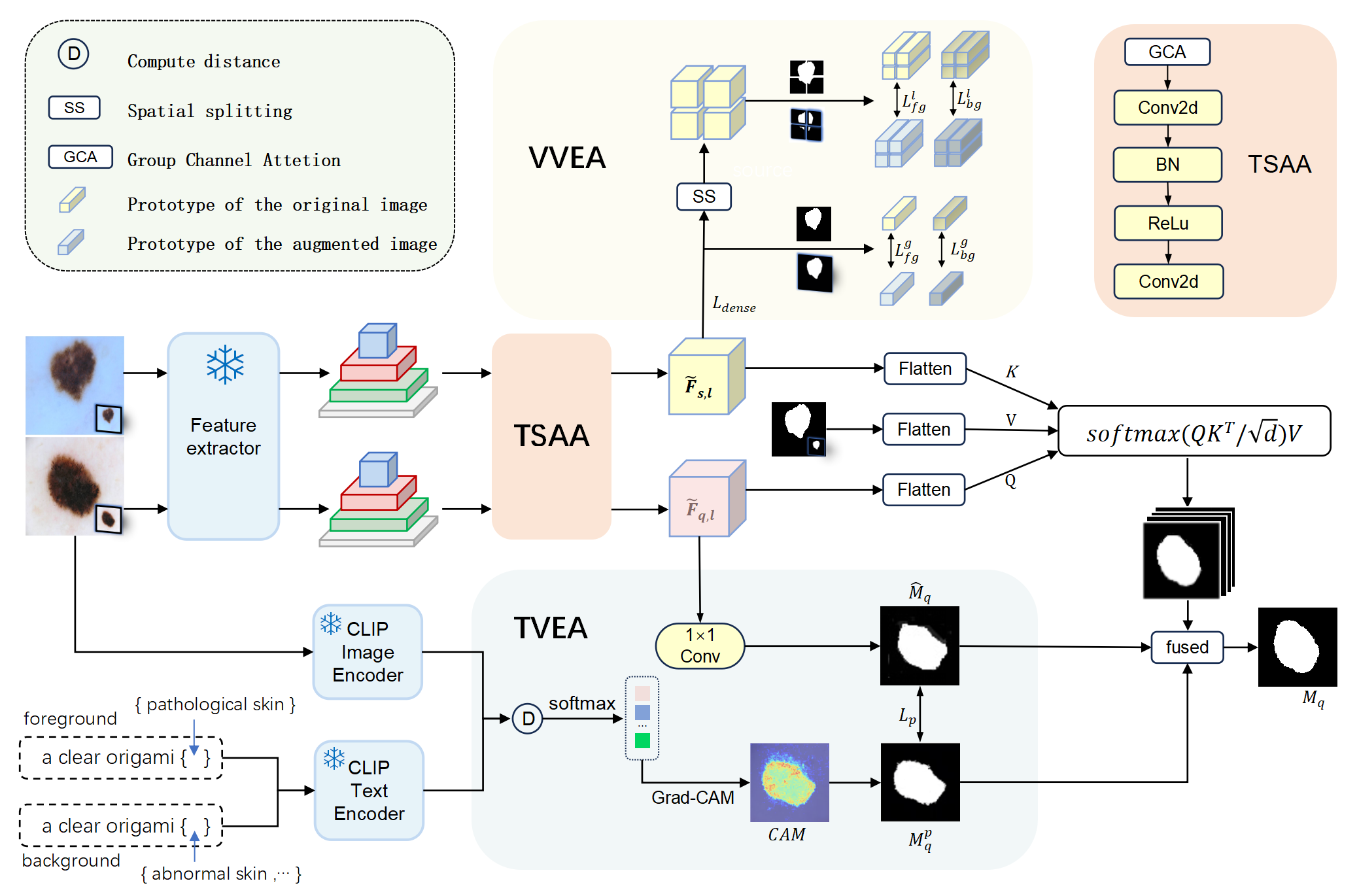}
    \caption{Overall framework of TVGTANet.}
    \Description{Overall framework of TVGTANet.}
    \label{fig:framework}
\end{figure*}

\section{Methodology}

\subsection{Problem Setting}

ABCDFSS~\cite{24} introduced a novel cross-domain few-shot segmentation (CD-FSS) task formulation. Its core idea is to achieve performance surpassing traditional methods without utilizing any source domain data. This formulation focuses solely on the target domain dataset $(X_t, Y_t)$, where $X_t$ denotes the input data distribution and $Y_t$ represents the label space. Under this formulation,  it is assumed that only a very limited number of annotated samples exist in the label space $Y_t$ of the target domain, and the model does not rely on any source domain data $(X_s, Y_s)$. 

\subsection{Overview}
The overall framework of TVGTANet is illustrated in Figure~\ref{fig:framework}. It consists of a Task-Specific Attention Adapter (TSAA) module, a Visual-Visual Embedding Alignment (VVEA) Module, and a Text-Visual Embedding Alignment (TVEA) Module. The TSAA module is designed to be appended to each intermediate layer of the backbone network, and achieves target task adaptation by integrating the Group Channel Attention (GCA) mechanism. During the training phase, both query and support images are augmented to obtain multiple views. The parameters of the TSAA module are optimized via the VVEA and TVEA modules. Specifically, the VVEA module enforces global and local cross-view class-agnostic embedding consistency, as well as within-support set class consistency. Concurrently, the TVEA module uses specially designed text prompts to guide the extraction of key semantics in target images and leverages a frozen CLIP model to directly generate pseudo segmentation masks for target domain images. These pseudo masks then serve as supervisory signals to train the pixel-level classifier. Throughout the training process, we freeze the parameters of both the backbone network and the CLIP network, updating only the additional task adaptation modules. 
During the prediction phase, given the task-adapted features, we obtain a coarse prediction map for each layer by calculating the similarity between query image pixels and support image foreground pixels. Simultaneously, we utilize the CLIP model and the pixel-level classifier to obtain corresponding prediction maps separately. Finally, we fuse the aforementioned prediction maps, perform thresholding, and refine them as needed to obtain the final segmentation mask.

\subsection{Task-Specific Attention Adapter (TSAA)}
To effectively enhance feature expressive capability and adapt to task requirements, we propose a Task-Specific Attention Adapter (TSAA) network that is appended to the backbone network. The TSAA network comprises a Group Channel Attention (GCA) layer and two convolutional layers.

\begin{figure}[t]
    \centering
    \includegraphics[width=\columnwidth]{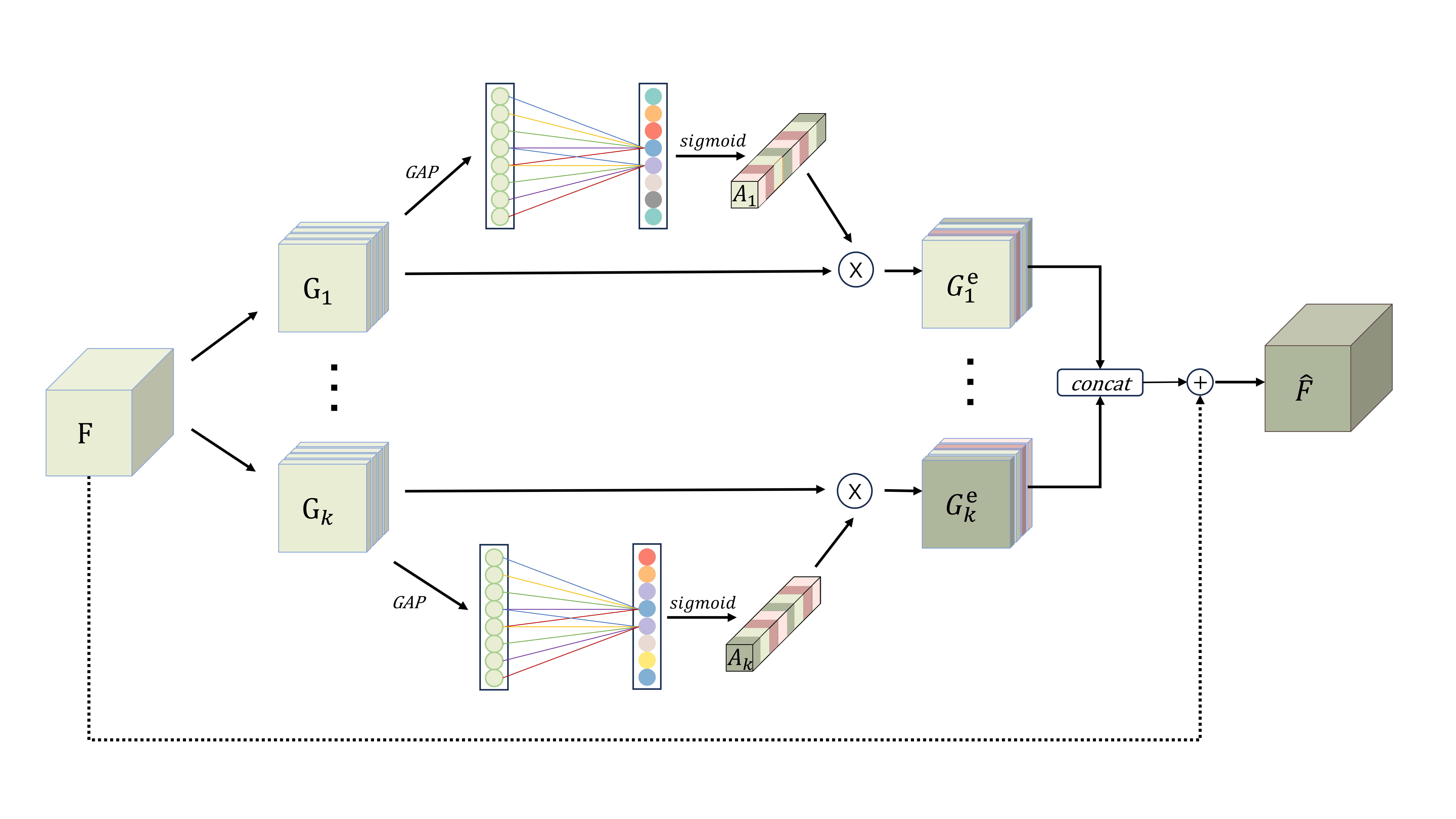}
    \vspace{-15pt}  
    \Description{The specific implementation of the Group Channel Attention(GCA)}
    \caption{The structure of the Group Channel Attention.}
    \label{fig:GCA}
\end{figure}

\textbf{Group Channel Attention (GCA).}
Prototypical computation often fails to adequately address intra-class variability. To mitigate this, we propose a feature enhancement method based on channel grouping and an attention mechanism, namely Group Channel Attention(GCA). Along the channel dimension, the image feature \( F \) is partitioned into  \( g \)-channel groups. For each group \( G_k \in \mathbb{R}^{g \times H \times W} \), a saliency weight \( A_k \) is computed via a channel attention mechanism:
\[
A_k = \sigma\Bigl(W_2\, \delta\bigl(W_1\, \mathrm{GAP}(G_k)\bigr)\Bigr) \tag{1}
\]
Here, \(\mathrm{GAP}\) denotes global average pooling which extracts global contextual information within the group channels. \(W_1\) and \(W_2\) are learnable weight matrices, while \(\delta\) and \(\sigma\) denote the ReLU and Sigmoid activation functions, respectively. Subsequently, the saliency weight \( A_k \) is applied to the feature group \( G_k \) to obtain the final enhanced feature \( \hat{G}_k \):
\[
\hat{G}_k = G_k \odot A_k \tag{2}
\]
In the above, \(\odot\) denotes element-wise multiplication performed channel-wise. The enhanced grouped features are then concatenated to form the locally enhanced feature \(F^e\).

\textbf{Task-Adaptive Feature Generation.}
After the GCA operation, the support feature \(F^e_s\) and the query feature \(F^e_q\), are input into convolutional blocks to generate task-adaptive features. The transformation operation of the convolutional block is expressed as:
\[
\tilde{F} = \operatorname{Conv}\Bigl(\operatorname{ReLU}\bigl(\operatorname{BN}(\operatorname{Conv}(F^e))\bigr)\Bigr) \tag{3}
\]
Here, \(\tilde{F}\) denotes the final task-adaptive feature. The feature \(\tilde{F}\) integrates global contextual information through dynamic adaptation. 

\subsection{Visual-Visual Embedding Alignment}
To train the appended TSAA module, both the query and support images are augmented to generate multiple views. Our aim is to identify task-relevant features by enforcing view-independent embedding consistency and global-local intra-class support consistency. We first achieve a degree of view-independent embedding consistency by computing a dense contrastive loss \(L_{\text{dense}}\) between the features extracted from augmented and non-augmented images through self-supervised learning, as described in ~\cite{24}. Furthermore, we propose a Global-Local Prototype Alignment strategy. Intuitively, the class prototypes across different views of the same scene should remain consistent given equivalent semantic information. Inspired by this, ABCDFSS ~\cite{24} previously proposed a class alignment method based on a global contrastive loss \(L_{\text{global}}\). However, the aforementioned method might lose local detailed information during the global prototype aggregation process. To address this limitation and enhance consistency, building upon the global prototypes, we further propose a class alignment method based on a local contrastive loss.
We perform feature segmentation on the spatial scale by dividing the task-adaptive support feature \(\tilde{F}_s\) into \( n\times n \) sub-blocks, with each sub-block representing the feature distribution within its local region, i.e., 
$\tilde{F}_s = \{F_s^{(i,j)} \mid i,j=1,\ldots,n\}$, $F_s^{(i,j)} \in \mathbb{R}^{\frac{H}{n} \times \frac{W}{n} \times C}$.
For each sub-block $F_s^{(i,j)}$, combined with its corresponding support image mask sub-block \( M_s^{(i,j)} \), we compute the local prototypes of the foreground and background:
\[
P_{fg}^{(i,j)} = \frac{\sum_{p \in F_s^{(i,j)}} m_p^s\, f_p}{\sum_{p \in F_s^{(i,j)}} m_p^s} \tag{4}
\]

\[
P_{bg}^{(i,j)} = \frac{\sum_{p \in F_s^{(i,j)}} (1 - m_p^s)\, f_p}{\sum_{p \in F_s^{(i,j)}} (1 - m_p^s)} \tag{5}
\]
where \(f_p\) denotes the feature vector at the given location $p$ and \(m_p^s\) denotes the corresponding foreground label in the mask. The same method is applied to the augmented image features \(\hat{F}_s\) to compute the corresponding augmented foreground prototype \(\hat{P}_{fg}^{(i,j)}\) and augmented background prototype \(\hat{P}_{bg}^{(i,j)}\). As illustrated in Figure~\ref{fig:framework}. Given the original feature \(P\) and the augmented feature \(\hat{P}\) for a given patch, we compute: (1) the similarity for the foreground prototype, \(\sin(P_{fg}^{(i,j)}, \hat{P}_{fg}^{(i,j)})\), and (2) the similarity between the foreground and background prototypes, \(\sin(P_{fg}^{(i,j)}, \hat{P}_{bg}^{(i,j)})\). Here, the similarity is measured by a symmetric Structural Similarity (SSIM) function~\cite{wang2004image}. The local contrast loss is then defined as:
\[
\begin{split}
L_{\text{local}}^{(i,j)} = \log\Biggl(1 + \exp\Bigl(&\sin\bigl(P_{fg}^{(i,j)},\, \hat{P}_{bg}^{(i,j)}\bigr)\\[0.5em]
&-\sin\bigl(P_{fg}^{(i,j)},\, \hat{P}_{fg}^{(i,j)}\bigr)\Bigr)\Biggr)
\end{split} \tag{6}
\]
The final local prototype loss \(L_{\text{local}}\) is obtained by averaging the loss over all enhanced samples:
\[
L_{\text{local}} = \frac{1}{n^2} \sum_{i=1}^{n} \sum_{j=1}^{n} L_{\text{local}}^{(i,j)} \tag{7}
\]

\subsection{Text-Visual Embedding Alignment}

In few-shot tasks, significant distribution shifts between source and target domains hinder traditional methods from achieving cross-domain semantic consistency. We propose a strategy that combines textual information with image-based pseudo-label generation to guide model training and enhance target performance.

We generate Class Activation Maps (CAMs) using image-level labels and improve pseudo-label quality via noise suppression and dynamic thresholding. A frozen CLIP image encoder extracts image features from the feature map prior to its final self-attention layer. Concurrently, text prompts, constructed from target foreground and background labels, are embedded via a frozen CLIP text encoder to preserve robust image-text alignment.

For pseudo-label generation, foreground text is set as the target object's name for single-label tasks (with background text formed by prefixing "non"), and for multi-label tasks, the image's class is used as foreground while background text is built by combining "non" with the names of remaining classes. Let \(s_c\) denote the normalized score for class \(c\). The softmax operation is used to amplify the target class and suppress the background. Then we use Grad-CAM~\cite{32} to compute the contribution of each feature channel:
\[
w_k^c = \frac{1}{Z} \sum_{i,j}\Biggl(\frac{\partial Y^c}{\partial A_{ij}^k}\cdot s_c(1-s_c)
+ \sum_{c'\neq c}\frac{\partial Y^{(c')}}{\partial A_{ij}^k}\cdot s_c(-s_{c'})\Biggr) \tag{8}
\]
where \(w_k^c\) is the weight of the \(k\)th feature map for class \(c\). These weights are applied and aggregated to produce the final CAM, which is then converted into a grayscale image via dynamic thresholding~\cite{otsu1975threshold} to serve as the pseudo-label \(M_q^p\).

As shown in Figure~\ref{fig:framework}, the TSAA module produces a rough query mask \(\hat{M}_q\), which may be inaccurate early on due to weak semantic cues. The CLIP-generated pseudo-label \(M_q^p\), combined with image semantics, refines foreground localization.

To ensure accurate foreground capture, we propose a self-supervised loss aligning \(\hat{M}_q\) with \(M_q^p\). The rough mask and pseudo-label are interpolated to the same scale, and the cross-entropy loss is computed as:
\[
L_{\text{pascal}} = \text{CrossEntropy}(\hat{M}_q, M_q^p) \tag{9}
\]
This loss enforces consistency between the rough mask and pseudo-label, guiding the model to learn more precise segmentation features and leveraging cross-modal contrast for improved adaptability.

For all layers in the pyramid, the TSAA module is then optimized on the combined loss \(\mathcal{L}\): 
\[
\mathcal{L} = \sum_{l=1}^{N} \Bigl( L_{\text{local}}^{l} + L_{\text{global}}^{l} + L_{\text{pascal}}^{l} + L_{\text{dense}}^{l} \Bigr) \tag{10}
\]
where \(N\) denotes the number of layers. By progressively fusing the local contrast, global representation, pseudo-label loss, and dense contrast loss across scales, the segmentation performance on multi-scale features is effectively enhanced.

\subsection{Segmentation}

After the TSAA module generates the task-adaptive features, we follow the multi-scale and multi-layer cross-attention mask aggregation method proposed in \cite{44} to generate the segmentation mask for the query feature. First, \(\tilde{F}_{q}\), \(\tilde{F}_{s}\), and the support mask \(M_s\) are unfolded into one-dimensional sequences at the pixel level. After adding positional encoding, they form the query matrix \(Q\), key matrix \(K\), and value matrix \(V\). The multi-head attention mechanism is then employed to compute the per-pixel similarity between the query feature and the support feature. The attention weights for the query pixels are obtained according to:
\[
\operatorname{Attn}(Q, K, V) = \operatorname{softmax}\Bigl(\frac{QK^\mathrm{T}}{\sqrt{d}}\Bigr)V \tag{11}
\]
where \(d\) is the scaling factor of the feature dimension. The computed attention weights aggregate the support mask for each query pixel, generating an attention query mask \(\overline{M}_q^i\) for each layer. Masks generated at each layer are concatenated to form a multi-layer mask set \(\overline{M}_q=\{\overline{M}_q^i,i=1,..,N\}\).

At last, we fuse the rough query mask \(\hat{M}_q\), the CLIP-generated pseudo-label \(M_q^p\), and the attention query masks \(\overline{M}_q\) through skip connections to produce the refined segmentation result.

\section{Experiments}
\begin{table*}[htbp]
    \centering
    \captionsetup{width=0.88\textwidth, justification=centering} 
    \resizebox{0.9\textwidth}{!}{
    \begin{tabular}{lcccccccccc}
        \hline
        Methods       & \multicolumn{2}{c}{DeepGlobe} & \multicolumn{2}{c}{ISIC2018} & \multicolumn{2}{c}{Chest X-ray} & \multicolumn{2}{c}{FSS-1000} & \multicolumn{2}{c}{Average} \\
        \hline
                      & 1-shot & 5-shot & 1-shot & 5-shot & 1-shot & 5-shot & 1-shot & 5-shot & 1-shot & 5-shot \\
        \hline
        PMNet ~\cite{33}    & 37.10  & 41.60  & 35.40  & 39.10  & 30.60  & 31.30  & 75.10  & 84.60  & 41.00  & 46.00  \\
        HDMNet ~\cite{34}   & 35.40  & 39.10  & 33.00  & 35.00  & 30.60  & 31.30  & 75.10  & 78.60  & 41.00  & 46.00  \\
        PATNet ~\cite{23}   & 37.89  & 42.97  & 41.16  & 53.58  & 66.61  & 70.08  & 78.59  & 81.23  & 56.02  & 61.99  \\
        RestNet ~\cite{35}  & 22.70  & 29.90  & 42.30  & 51.10  & 73.70  & 73.70  & 74.00  & 74.00  & 53.42  & 59.90  \\
        ABCDFSS ~\cite{24}  & 42.60  & 49.00  & 45.70  & 53.30  & 79.80  & 81.40  & 74.00  & 76.00  & 60.70  & 65.00  \\
        DRANet ~\cite{21}   & 39.50  & 50.12  & 40.77  & 48.78  & 82.35  & 82.12  & 79.50  & 81.40  & 60.86  & 65.42  \\
        APSeg ~\cite{19}   & 35.94  & 39.98  & 45.43  & 53.98  & 84.10  & 84.50  & \textbf{79.71}   & \textbf{81.94}   & 61.30   & 65.09   \\
        Our           &  \textbf{42.04}  & \textbf{50.67}   & \textbf{47.21 }  & \textbf{58.75}   & \textbf{84.58}   & \textbf{87.27}   & 78.32  & 81.44  & \textbf{63.04}  & \textbf{69.53}  \\
        \hline
    \end{tabular}}
    \caption{Comparison of mIoU results on four datasets for cross-domain few-shot methods under 1-shot and 5-shot settings. Bold indicates the best performance for each method. Note: PMNet [4] did not report results on the ISIC dataset.}
    \label{tab:performance}
\end{table*}
\subsection{Datasets}
We utilize the same target domain datasets as \cite{23,19,20}, but unlike these approaches, we do not employ any source domain datasets. The target domain datasets include DeepGlobe~\cite{39}, FSS-1000~\cite{43}, ISIC2018~\cite{40}, and Chest X-ray~\cite{41,42}. These four datasets exhibit significant differences in both content domains and image styles, highlighting the inherent challenges in cross-domain tasks. The FSS-1000 dataset covers 1000 categories of everyday objects, with only 10 images per category. The DeepGlobe dataset focuses on remote sensing satellite imagery and comprises complex land surface scenes. The ISIC2018 dataset contains medical images of skin lesions, concentrating on the segmentation of benign and malignant lesions. Finally, the Chest X-ray dataset is used for lung region segmentation, where the medical imaging style differs markedly from that of natural images.

\begin{figure}[t]
    \centering
    \includegraphics[width=\columnwidth]{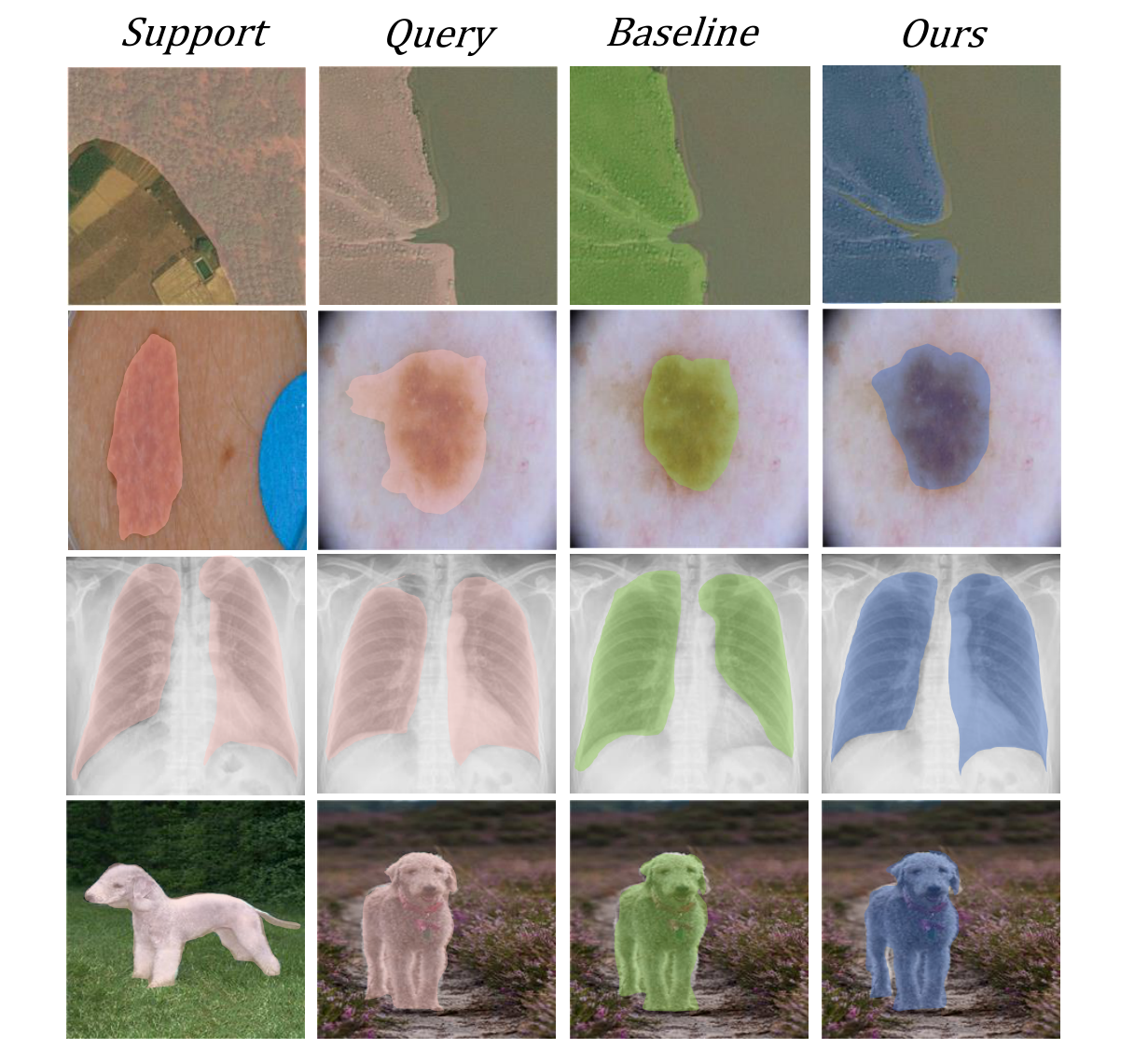}
    \vspace{-15pt} 
    \Description{Visualization results of our method on CD-FSS for 1-way 1-shot segmentation. The labels of the support and query images are shown in pink, while the predictions of the baseline and our method are displayed in green and blue, respectively. We use ABCDFSS~\cite{24} as our baseline.}
    \Description{Visualization results of our method on CD-FSS for 1-way 1-shot segmentation. The labels of the support and query images are shown in pink, while the predictions of the baseline and our method are displayed in green and blue, respectively.}
    \caption{Visualization results of our method on CD-FSS for 1-way 1-shot segmentation. The labels of the support and query images are shown in pink, while the predictions of the baseline and our method are displayed in green and blue, respectively. We use ABCDFSS~\cite{24} as our baseline.}
    \Description{Visualization results of our method on CD-FSS for 1-way 1-shot segmentation. The labels of the support and query images are shown in pink, while the predictions of the baseline and our method are displayed in green and blue, respectively.}    
    \label{fig:figure4}
\end{figure}

\subsection{Experimental Details}
We utilize ResNet50~\cite{37} with pretrained weights on ImageNet~\cite{38} as the backbone for multi-scale feature extraction, with its parameters kept frozen during training. The image augmentation strategy is based on affine transformations, and both support and query images are resized to \(400 \times 400\). In the GCA module, the channel projection dimension is set to 16, and the number of sub-blocks for local features is configured to 4. The training process utilizes the SGD optimizer for a total of 25 epochs, with a learning rate of \(1 \times 10^{-2}\). For each data category, task-specific adaptation optimization is performed only in the first episode; subsequent episodes directly use the optimized weights obtained from the first episode. In the Text-Visual Embedding Alignment Module, we use a frozen pretrained CLIP model (with ViT-B/16 as the encoder) and define the target class as the foreground, while treating non-target classes as background, in order to reduce confusion between foreground and background. In addition, the generation quality of pseudo-labels is optimized via dynamic threshold adjustment. Finally, the segmentation masks are generated by post-processing with dense CRF~\cite{36}. Performance is evaluated using mean Intersection over Union (mIoU) and FB-IoU.

\begin{table*}[htbp]
    \centering
    \begin{tabular}{lcccccccccc}
        \hline
        Methods       & \multicolumn{2}{c}{DeepGlobe} & \multicolumn{2}{c}{ISIC2018} & \multicolumn{2}{c}{Chest X-ray} & \multicolumn{2}{c}{FSS-1000} & \multicolumn{2}{c}{Average} \\
        \hline
                      & 1-shot & 5-shot & 1-shot & 5-shot & 1-shot & 5-shot & 1-shot & 5-shot & 1-shot & 5-shot \\
        \hline
        PATNet~\cite{23}    & 45.7   & 50.2   & \textbf{62.1}   & 68.8   & 72.6   & 73.3   & 85.5   & 87.2   & 66.5   & 69.9   \\
        ABCDFSS~\cite{24}   & \textbf{47.7}   & 54.6   & 60.3   & 66.1   & 86.1   & 87.3   & 82.7   & 84.2   & 69.2   & 73.1   \\
        Our           & 46.5   & \textbf{54.7}   & 61.1   & \textbf{72.6}   & \textbf{89.6}   & \textbf{91.4}   & \textbf{85.5}   & \textbf{88.2}   & \textbf{70.7}   & \textbf{76.7}   \\
        \hline
    \end{tabular}
    \caption{Additional evaluation using FB-IoU as an evaluation metric.}
    \label{tab:comparison}
\end{table*}
\subsection{Comparison with Existing Methods}
\hyperref[tab:performance]{Table~1} presents a comparison between our proposed method and several state-of-the-art cross-domain few-shot segmentation (CD-FSS) approaches. Although our method does not utilize source domain datasets, it demonstrates significant superiority across all datasets. Averaged over the four datasets, our method outperforms the previously best-performing DRA by 2.18\% and 4.11\% in the 1-shot and 5-shot settings, respectively.

\begin{figure}[t]
    \centering
    \includegraphics[width=\columnwidth]{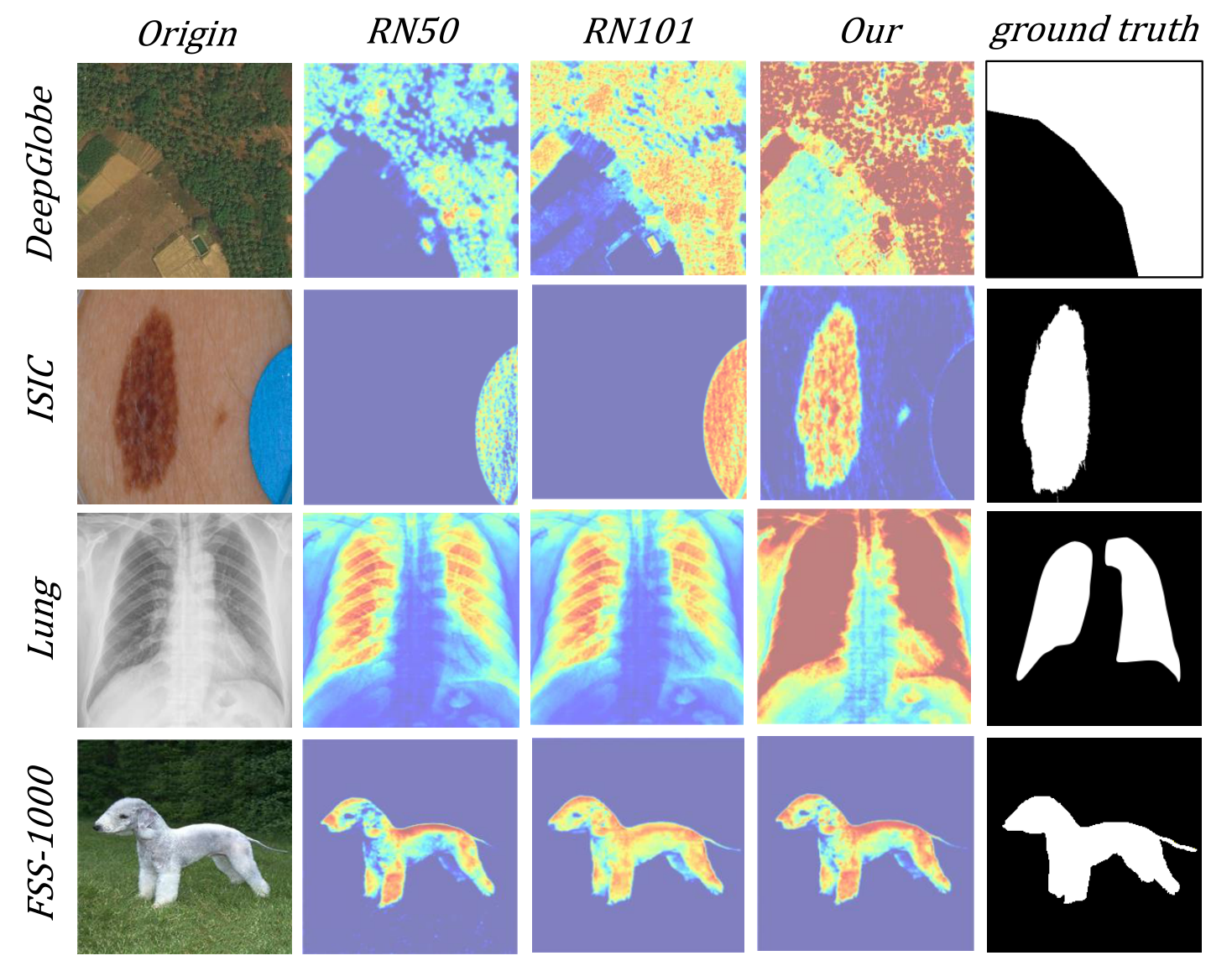}
    \vspace{-15pt} 
    \Description{The images show CAM maps generated by different CLIP models on four distinct datasets. In the heatmaps, highly concentrated bright regions typically indicate the model's focus on the target foreground, while darker areas represent responses to the background.}
    \caption{CAM visualizations of different models. The images show CAM maps generated by different CLIP models on four distinct datasets. In the heatmaps, highly concentrated bright regions typically indicate the model's focus on the target foreground, while darker areas represent responses to the background.}
    \label{fig:CAM}
\end{figure}

On the ISIC2018 dataset, our method effectively distinguishes between skin and skin-adherent regions through the TVEA module (as shown in Figure~\ref{fig:figure4} and Figure~\ref{fig:CAM}). This enhances intra-class feature aggregation and reduces intra-class variability. Under the 1-shot and 5-shot settings, our method improves upon the current best method by 1.78\% and 4.77\%, respectively. These results highlight the suitability of the pseudo-labeling strategy for this task.

For the Chest X-ray dataset, the foreground-background discrimination provided by the pseudo-labeling method effectively separates bony structures from lung regions (as depicted in the Lung part of Figure~\ref{fig:CAM}). This significantly reduces the misclassification of bone regions, which has been observed in previous approaches. Although earlier methods have achieved high accuracy on this dataset, our method further refines local details and enhances overall performance. These results demonstrate the capability of VVEA in refining local boundaries and edges. Specifically, our method outperforms ABCDFSS~\cite{24} by 4.78\% and 5.87\% in the 1-shot and 5-shot settings, respectively.

For the FSS-1000 dataset, the segmentation network decoders in PATNet and PMNet progressively recover spatial resolution through upsampling and multi-layer convolutions. This helps them accurately reconstruct the target boundaries and spatial details. As our model does not incorporate a decoder module, its ability to capture boundary details and spatial resolution is somewhat limited. Although our performance lags behind PMNet—the best performer on this dataset—it remains comparable to APSeg and DRA.

In the DeepGlobe dataset, although ABCDFSS achieves the highest accuracy under the 1-shot setting, our method shows better performance in the 5-shot setting. According to ABCDFSS, annotation quality issues in this dataset, along with ambiguous boundaries between the pasture and forest categories, can mislead pseudo-label generation and hinder model optimization.

We additionally employ FB-IoU as an evaluation metric to better assess the quality of foreground and background segmentation achieved by our method. This metric is particularly useful for small lesion areas in the ISIC dataset, where lesions are irregular in shape and occupy only a small portion of the image. In such cases, FB-IoU more accurately reflects the effectiveness of the proposed method.

Figure~\ref{fig:CAM} illustrates that, from left to right, as the model is progressively refined, the generated CAM maps show increasingly detailed and focused responses over the target regions. However, this improvement comes at the cost of increased computational overhead. Under limited hardware resources, TVEA can be executed independently, and its outputs can be passed into VVEA. This pipeline effectively reduces GPU memory usage. When using the RN50 variant of CLIP to generate CAM maps, the model successfully identifies the \textit{bedlington\_terrier} in the FSS-1000 datasets. However, misclassifications can occur when processing bony structures in the Lung dataset and specific skin patches in the ISIC dataset. These issues are significantly mitigated when using ViT-B/16. To further mitigate the label noise in the pseudo mask, we further use DenseCRF for post-processing after applying the dynamic threshold strategy. See appendix for detailed results.

\subsection{Ablation Study}

We validate the effectiveness of the proposed VVEA and TVEA by evaluating the accuracy under both 1-shot and 5-shot settings across the four datasets. \hyperref[tab:fb_iou]{Table~3} presents the impact of each individual component on model performance. To establish a baseline, we removed VVEA and directly used the frozen feature extraction network to extract features from images. A \(1 \times 1\) convolution kernel was then applied to transform the features across different channel scales. Subsequently, multi-layer mask sets were obtained through dense comparisons between the support and query features. We also removed TVEA module, meaning that pseudo-labels and rough query masks were not incorporated into the multi-layer mask set after the dense comparison.

\begin{table}[htbp]
    \centering
    \begin{tabular}{ccccc}
        \hline
        VVEA & TVEA & 1-shot & 5-shot \\
        \hline
                &      & 50.70  & 53.64  \\
        \checkmark &              & 61.66  & 67.75  \\
         &   \checkmark            & 59.61  & --     \\
        \checkmark & \checkmark   & \textbf{63.04}  & \textbf{69.53}  \\
        \hline
    \end{tabular}
    \caption{Ablation Study of Key Components of TVGTANet on CD-FSS (Results from ISIC)}
    \label{tab:fb_iou}
\end{table}

\begin{figure}[t]
    \centering
    \includegraphics[width=\columnwidth]{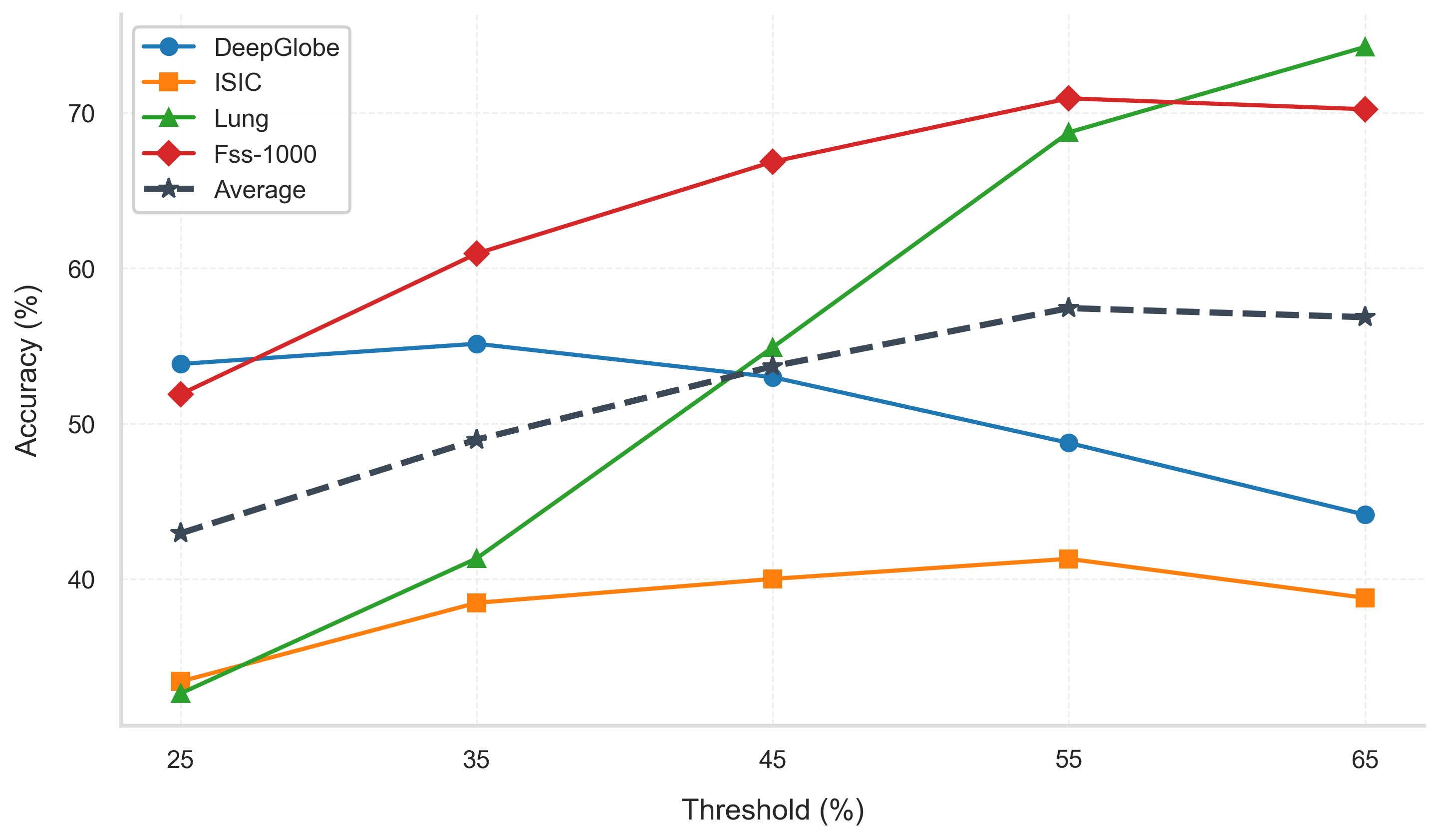}    
    \vspace{-15pt}  
    \Description{Accuracy of Different Datasets under Various Thresholds in the TVEA Module.}
    \caption{Accuracy of Different Datasets under Various Thresholds in the TVEA Module.}
    \label{fig:plot}
\end{figure}

The results in \hyperref[tab:fb_iou]{Table~3} clearly show that incorporating both VVEA and TVEA improves the model performance. Specifically, VVEA enables both local and global task adaptation, significantly boosting segmentation accuracy. The 1-shot and 5-shot accuracies increase to 61.60\% and 67.75\%, respectively.

Although 5-shot data is unavailable for TVEA, we evaluated its performance on each of the four datasets using the standard mIoU metric for semantic segmentation. Since TVEA is a weakly supervised module that does not utilize ground truth masks, we further analyzed how different thresholds impact pseudo-label generation, as shown in Figure~\ref{fig:plot}. The best-performing threshold was selected, and the corresponding results are reported in \hyperref[tab:fb_iou]{Table~3} for the TVEA-only configuration. By combining both VVEA and TVEA modules, we achieves state-of-the-art segmentation results, reaching 63.04\% and 69.53\% in the 1-shot and 5-shot settings, respectively.
\begin{table}[htbp]
    \footnotesize
    \setlength{\tabcolsep}{4pt}
    \centering
    \begin{tabular}{c cc cc cc cc}
        \hline
        Channels  & \multicolumn{2}{c}{0} & \multicolumn{2}{c}{8} & \multicolumn{2}{c}{16} & \multicolumn{2}{c}{32} \\
        \hline
                  & 1-shot& 5-shot& 1-shot& 5-shot& 1-shot& 5-shot& 1-shot& 5-shot\\
        \hline
        mIoU      & 61.32& 67.65& 62.85& 68.96& \textbf{63.00}& \textbf{69.50}& 62.83& 69.08\\
        FB-IoU    & 68.18& 73.25& 70.17& 75.86& \textbf{70.70}& \textbf{76.70}& 70.12& 75.97\\
        \hline
    \end{tabular}
    \caption{Impact of Different Group Channel Numbers in Group Channel Attention (GCA)}
    \label{tab:plain}
\end{table}

\begin{table}
    \centering
    \begin{tabular}{c c}
        \hline
        Channels  & Learnable Params\\
        \hline
        0      &    1.05 M \\
        8      &    1.16 M \\
        16     &    1.21 M \\
        32     &    1.48 M \\
        128    &    2.76 M \\
        \hline
    \end{tabular}
    \caption{The Number of Parameters for Different Channel Grouping Numbers}
    \label{tab:plain2}
\end{table}

\textbf{Number of Channel Groups in GCA.} In the GCA module, multi-layer pyramid features are grouped along the channel dimension. A channel attention mechanism is then applied to reweight each group before recombining them to obtain enhanced channel-wise features. As shown in \hyperref[tab:plain]{Table~4}, the best performance is achieved when the number of channel groups is set to 16. This setting yields superior results in both 1-shot and 5-shot tasks under mIoU and FB-IoU metrics. As shown in \hyperref[tab:plain2]{Table~5}, increasing the number of channel groups significantly raises the number of learnable parameters. The 16-group setting strikes a favorable balance between performance and computational cost. These results demonstrate the effectiveness of channel-wise feature enhancement for cross-domain few-shot segmentation.

\begin{table}[ht]
\centering
\begin{tabular}{c|cc|cc}
\hline
\textbf{Num} & \multicolumn{2}{c|}{\textbf{mIoU}} & \multicolumn{2}{c}{\textbf{FB-IoU}} \\
            & 1-shot & 5-shot & 1-shot & 5-shot \\
\hline
0   & 61.41 & 68.33 & 67.74 & 75.06 \\
4   & \textbf{63.04} & \textbf{69.53} & \textbf{70.74} & \textbf{76.79} \\
9   & 62.66 & 69.10 & 69.85 & 76.18 \\
16  & 62.21 & 68.72 & 69.07 & 75.73 \\
25  & 61.56 & 68.05 & 68.25 & 75.24 \\
\hline
\end{tabular}
\caption{Impact of Different Feature Segmentation Block Numbers}
\label{tab:aux_samples}
\end{table}

\textbf{Number of Feature Segmentation Blocks.} The number of feature segmentation blocks is critical for computing the local prototype loss. It determines the granularity of feature division, thereby influencing the model's ability to capture fine details and small object boundaries. More segmentation blocks lead to finer resolution, but excessive segmentation can overfit to local noise. As shown in \hyperref[tab:aux_samples]{Table~6}, the optimal number is 4, which achieves the best balance and improves local feature contrast.
\subsection{Computational Efficiency}
To assess the computational efficiency of our proposed TVGTANet model, we conducted a comparative analysis against established CD-FSS models. This evaluation considered key performance metrics: inference time, inference memory, training time, training memory, and the number of trainable parameters. All experiments were performed using the ISIC2018 dataset on an RTX 4090 GPU, with a consistent batch size of 8. To optimize memory usage and standardize the training process, all input images were resized to 400x400 pixels. Single-image inference times were calculated as an average over five runs. As detailed in Table 7, our method demonstrates the most efficient training duration. Furthermore, its inference time is highly competitive, closely matching that of ABCDFSS [11], another leading source-free CD-FSS approach.
\begin{table}[ht]
    \footnotesize
    \setlength{\tabcolsep}{3.5pt}
    \centering       \label{tab:isic2018_performance_simple}
    \small
    \begin{tabular}{@{}lp{1.1cm}p{0.95cm}p{1.3cm}p{1.3cm}p{0.95cm}@{}}
        \toprule
        Method &Inf Time&Inf Mem&Train Time&Train Mem &Param \\
        \midrule
        PATNet~\cite{23}  & 50.41ms& 724 MB& 123m 07s& 9469MB  & 2.58M   \\
        DRANet~\cite{21}  & 48.72ms& 896MB& 94m25s & 14887 MB & 59.25M   \\
        
        ABCDFSS~\cite{24} & 24.83ms& 806MB& 2m17s& 1497MB  & 1.08M     \\
        Ours    & 32.46ms& 844MB& 1m27s  & 1432MB  & 1.21M   \\
        \bottomrule
    \end{tabular}
    \caption{Computational Efficiency Comparison on ISIC2018 (Simplified Headers)}
\end{table}

\section{Conclusion}
In this paper, we propose TVGTANet, a novel source-free CD-FSS method effectively leveraging visual and textual information. By introducing a Task-Specific Attention Adapter and employing cross-view and text-guided alignment, our approach achieves robust target domain adaptation without relying on source data. Experiments demonstrate TVGTANet's superior performance, highlighting the potential of text-guided learning in challenging cross-domain few-shot segmentation scenarios.

\section*{Acknowledgment}
This work was financially supported by the National Natural Science Foundation of China (Grant No.62266022), Natural Science Foundation of Jiangxi, China (Grant No.20242BAB25110). 
\bibliographystyle{ACM-Reference-Format}
\balance
\bibliography{refer} 










\end{document}